# Object-Centric Photometric Bundle Adjustment with Deep Shape Prior

Rui Zhu, Chaoyang Wang, Chen-Hsuan Lin, Ziyan Wang, Simon Lucey
The Robotics Institute, Carnegie Mellon University
{rz1, chaoyanw, chenhsul, ziyanw1}@andrew.cmu.edu, slucey@cs.cmu.edu

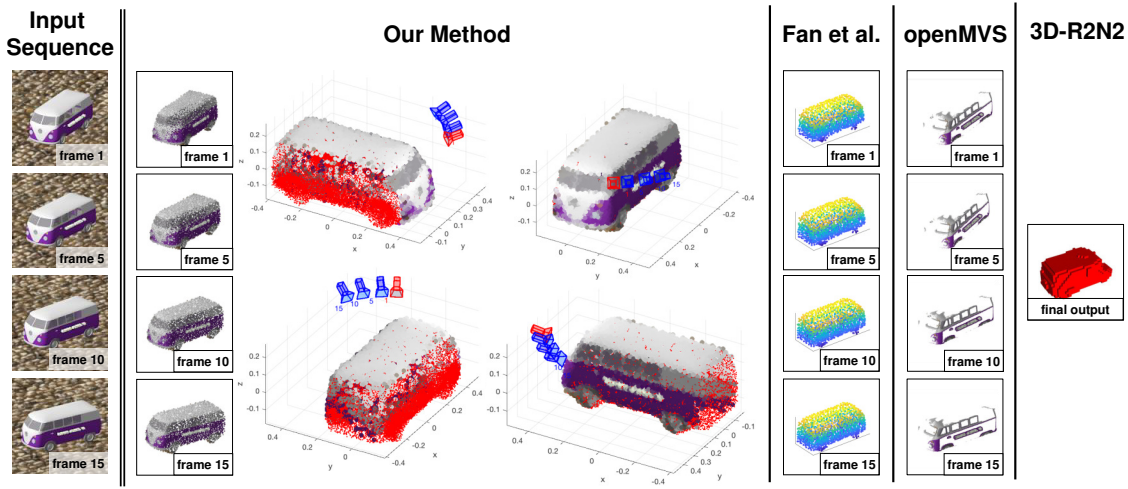

Figure 1: We introduce our approach which recovers both global camera poses and point-cloud-based shape in full 3D from a few observations in limited views, and yet guarantees photometric consistency across views. The subset of the reconstructed point cloud which are occluded from any views in the frame is marked red. In contrast Fan et al.[7] and 3D-R2N2[5] reconstructs shapes from single frames in canonical pose with no cameras, and openMVS[9] produces SfM results with accurate but sparse reconstruction for visible parts only.

## Abstract

*Reconstructing 3D shapes from a sequence of images has long been a problem of interest in computer vision. Classical Structure from Motion (SfM) methods have attempted to solve this problem through projected point displacement & bundle adjustment. More recently, deep methods have attempted to solve this problem by directly learning a relationship between geometry and appearance. There is, however, a significant gap between these two strategies. SfM tackles the problem from purely a geometric perspective, taking no account of the object shape prior. Modern deep methods more often throw away geometric constraints altogether, rendering the results unreliable. In this paper we make an effort to bring these two seemingly disparate strategies together. We introduce learned shape prior in the form of deep shape generators into Photometric Bundle Adjustment (PBA) and propose to accommodate full 3D shape generated by the shape prior within the optimization-based inference framework, demonstrating impressive results.*

## 1. Introduction

Dating from the early age of computer vision, most of the classic methods for reconstructing objects in 3D (*e.g.* multi-view stereo and structure from motion) have been approaching this problem purely based on multi-view geometry: they recover the 3D shape of an object from a number of densely sampled views by utilizing visual cues, including feature correspondence [11, 12, 2], photometric consistency [6, 3, 19] and silhouettes constraint [24, 10]. Although these methods are still among the state-of-the-art, they have been facing inherent limitations such as their inefficiency to deal with specularity or lack of texture, and usually rely heavily on a complete coverage of views to acquire full 3D dense reconstruction. To circumvent these issues, the vision community is exploring the possibility of introducing shape priors to the reconstruction pipeline.

The shape prior can be parameterized in the form of explicit rigid models, such as shape primitives [21, 15, 14], skeleton models [20] and meshes [13]. The 3D shape is then recovered by deforming or blending the models such that the key points/contours of the output shape are aligned to the 2D landmarks/silhouette on the images. With constraint from explicit shape priors, it is now possible to perform dense full 3D reconstruction from a few or even single images. However, for a shape collection with large intra-class variation, *e.g.* models of a category from the ShapeNet [4] repository, it is generally difficult to either retrieve from existing models [27], or to faithfully reconstruct the target object with deformation or blending, considering global correspondence between models needed for the blending is difficult to establish [13].

On the other hand, some recent works [7, 16, 28, 29, 8, 26, 25] choose to model the shape prior implicitly as a deep network, and train the network to output target representations (depth map [16], volumetric grid [28, 8, 29, 26, 25], or point cloud [7]) directly from a single image. These approaches are purely data driven, thus hold the promise of covering a large variety of 3D shapes given a high-capacity deep network. However, these methods are largely restricted to partial (single) view reconstruction, where the quality of the reconstruction is often impacted by self-occlusion or style ambiguity [5].

To take a step further, 3D-R2N2 [5] demonstrates that, by taking observation from multiple views, a recurrent network is able to refine reconstruction overtime. However, this method models the reconstruction process as a black box, giving away the essential constraint from multi-view geometry, such as photometric consistency across different views, which serves as a golden rule to measure the accuracy of the reconstruction. Although some methods [28, 29, 16] indeed use multi-view geometry in training time, no geometric constraints is explicitly enforced during one-pass inference. As a result, they are not able to optimize for or evaluate on those geometric metrics. This is in fact a severe drawback compared to geometry-based approach like the emerging technique of photometric bundle adjustment (PBA) [3].

In PBA, dense pixel-wise correspondence is established with the estimated inverse depth map and camera poses, injecting a natural yet strong geometric constraint into the system. However this photometric consistency breaks on textureless regions, so it tends to produce confident results only on edges with strong image gradients, inferior to the deep methods which are equipped with dense shape generators.

Hitherto, the central problem to address in our work is, can we combine both the goodness from the shape prior offered by deep networks, and multi-view geometric constraints to improve results of PBA. In this paper, we propose to use a deep point cloud generator derived from Fan *et al.* [7] to model the shape prior. The resulting point cloud is thus parameterized by a low dimensional style vector. In inference, we adopt a PBA-like procedure which jointly optimizes style and the camera poses to minimize the photometric consistency cost across multiple views. With this approach, we are able to inference a dense point cloud reconstruction from a few shots of the object, and yet sticks to multi-view geometry.

We summarize our contribution as follows:

- Unlike traditional photometric bundle adjustment which models point cloud as a set of independent points, we use a deep generative network to associate point cloud with a low dimensional shape style vector. This dependency between points provides prior knowledge about the structure of the shapes, and allows differentiable bidirectional mapping between shape and style embedding. As a result, the deep shape prior reduces the degrees of freedom (DoF) for optimization, and allows full 3D dense reconstruction from very few shots of the object.

- Compared to previous deep methods of shape from image(s), we are the first to explicitly enforce multi-view geometry constraints in inference. Therefore, our method is able to provide accurate camera pose guaranteed by the geometric consistency, which is either unavailable [7, 16, 28, 8, 26, 25] or unreliable [22, 29] in previous methods.

- We demonstrate that a RGB frame of an object taken from specific camera pose can be chained back to the full 3D shape with a differentiable operator, which enables gradient-descent-based PBA over camera poses and object style during inference with full shape.

### 1.1. Notation

Vectors are represented with lower-case bold font (*e.g.* $\mathbf{a}$). Matrices are in upper-case bold (*e.g.* $\mathbf{M}$) while scalars are low-cased (*e.g.* $a$ or $A$). Italicized upper-cased characters represent sets (*e.g.* $S$)

To denote the $l^{th}$ sample in a set (*e.g.* images, shapes), we use subscript (*e.g.* $\mathbf{M}_l$). Calligraphic symbols (*e.g.* $\pi(\mathbf{x}; \mathbf{p})$) denote functions which take in a vector or a scalar (*e.g.* $\mathbf{x}$) and meanwhile are parameterized by some parameters (*e.g.* $\mathbf{p}$).

## 2. Approach

### 2.1. Overview

In this paper, given a RGB sequence of a camera moving around a rigid object within limited baseline, we attempt to recover the full 3D shape of the object, as well as the camera extrinsics of each frame. Our objective is most close to that of PBA [19, 3], however instead of trying to recover the inverse depths associated with each pixel in the reference frame, we seek to reconstruct full 3D shape in the form of point cloud, untied from the pixels.

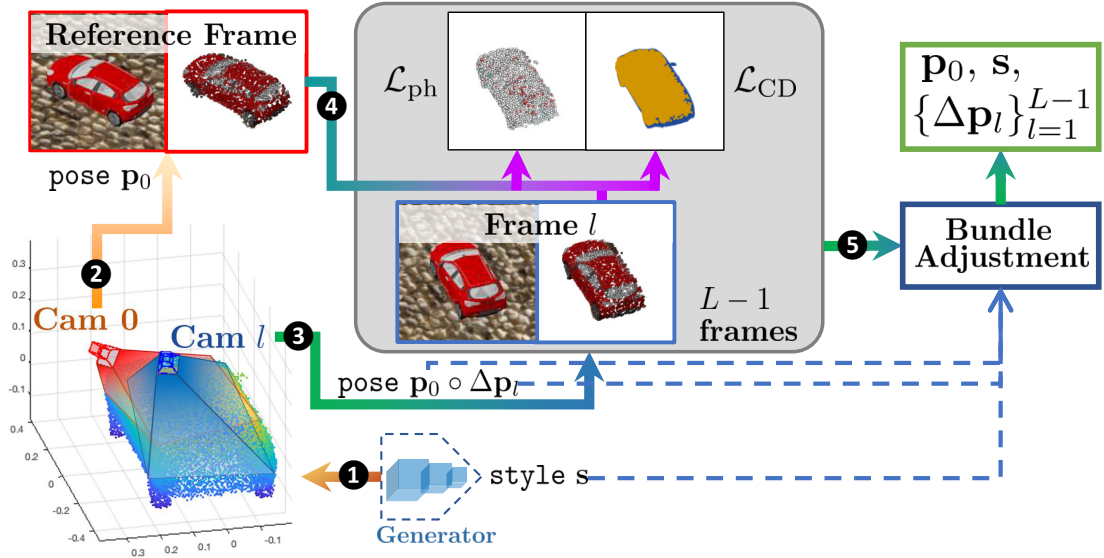

Figure 2: **Pipeline of the optimization.** (1) Generate shape $X$ with initialized style $\mathbf{s}$. (2) Get $X_{\mathbf{p}_0}$ with the pseudo-renderer and reproject it with initialized $\mathbf{p}_0$ to resample the reference frame. (3) Reproject $X_{\mathbf{p}_0}$ with $\mathbf{p}_0 \circ \mathbf{\Delta p}_l$ to resample the target frame $l$. (4) Compute the photometric loss $\mathcal{L}_{\text{ph}}$ and the silhouette loss $\mathcal{L}_{\text{CD}}$. (5) Bundle adjust $\mathbf{p}_0$, $\mathbf{s}$ and $\mathbf{\Delta p}_l$ by minimizing $\mathcal{L}_{\text{ph}}$ or $\mathcal{L}_{\text{ph}} + \lambda \mathcal{L}_{\text{CD}}$. Note that in a scenario where we know the silhouettes of the object inside the frames, the final objective becomes $\mathcal{L} = \mathcal{L}_{\text{ph}} + \lambda \mathcal{L}_{\text{CD}}$ as introduced in Section 2.3, and accordingly the computation of 2D silhouette loss is needed.

We experiment in controlled conditions, where image sequences are synthesized with a rendering pipeline from [22], hence we are equipped with the ground truth style and the camera poses for later use.

The shape is represented as point cloud generated from an trained shape autoencoder for embedding a repository of CAD models, where the generator of the autoencoder serves as a differentiable function mapping from a latent vector indicating object style to a point cloud. After the autoencoder is trained, a style regressor and a pose regressor are trained to regress from synthetic images with style and pose variations to their individual ground truth style vectors and pose parameters, similar to Zhu et al. [29]. This concludes all the network training work in our method.

In inference, the overview of our pipeline is shown in Fig. 2. Given an input image sequence $\{\mathcal{I}_l\}_{l=0}^{L-1}$ of an object transformed overtime, we treat the first frame (frame 0) as the reference (or template/source) frame, and attempt to align all the subsequent $L-1$ target frames to it. To achieve this, we optimize for three set of variables: the unique style vector $\mathbf{s}$ for the rigid object, the global [1] camera pose $\mathbf{p_0}$ of the reference frame, as well as the relative camera motion parameters from the reference frame to the target frames $\{\mathbf{\Delta p}_l\}_{l=1}^{L-1}$. These variables are initialized on the first step of initialization by running the trained regressors on the reference frame[2].

We will refer to the camera extrinsics as **pose** for short in the context. In our paper, for a RGB image $\mathcal{I}$ of an object with constant style taken from pose $\mathbf{p}$ in world frame, we represent the image as a function of subpixel locations $\mathbf{u} = [u, v]^T$, and parameterized by a style parameter $\mathbf{s}$ and pose parameter $\mathbf{p}$, that is, $\mathcal{I}(\mathbf{u}; \mathbf{p}, \mathbf{s}) : \mathbb{R}^2 \rightarrow \mathbb{R}^3$. For a point cloud with $N$ points, we use a collection of 3D point positions $X := \{\mathbf{x}_i\}_{i=1}^{N}$, where $\mathbf{x}_i \in \mathbb{R}^3$.

### 2.2. Shape Prior

We parameterize the point cloud with a shape prior. The shape prior $\mathcal{G}(\mathbf{s}) = \{\mathbf{x}_i\}_{i=1}^{N}$ is a differentiable deep generator function outputting a point cloud from a style embedding vector. The shape generator serves as a non-rigid shape prior over the style space. It embeds the entire set of points into a latent space, thus a forward pass gives the positions of every single point, and a backward propagation passes the gradients of the entire set or a subset of the points all the way back to the style vector.

The benefit of this parameterization is, 1) it reconstructs full 3D point clouds from a vector, invariant to viewpoints;

---

[1] Considering we are playing with a CAD repository of aligned meshes, we define the world frame as one aligned to the axes chosen by the dataset.

[2] We also tested with another initiation strategy for style using median/mean estimation from all frames, no significant difference is observed.

2) it embeds a potentially large set of dense points with DoF $3N$ into a latent space with relatively fewer dimensions, reducing the DoF of the optimization problem, yet maintains a differentiable one-to-one mapping between the style embedding and point set; 3) spatial constraints are implicitly imposed by the convolutional generator [7] so that the generated points are glued with neighbours, hence more robust to noise; 4) gradients of a subset of the generated point cloud can be back propagated to the style vector, changing the global shape of the output. We will cover the last point in detail when we describe the optimization process in Section 2.3.

### 2.3. Inference of Poses and Style Through Joint Optimization

**Camera Model**  We follow [29] by utilizing exponential twist for parameterizing the camera pose in world frame as well as the relative camera motions between the reference frame and target frames. Assuming a camera with known intrinsic matrix $\mathbf{K}$, the projection of an point $\mathbf{x}_i \in \{\mathbf{x}_i\}_{i=1}^N$ under the extrinsic parameters can be written as:

$$\pi(\mathbf{x}_i; \mathbf{p}) = \mathbf{K}(\mathbf{R}\mathbf{x}_i + \mathbf{t}) \in \mathbb{R}^2, \; i \in [1, N] \quad (1)$$

where for a rotation around unit axis $\mathbf{n} = [n_1, n_2, n_3]^T$ for radian $\phi$, the rotation matrix is given by $\mathbf{R} = \exp([\mathbf{n}]\hat{\,}\phi) \in \mathbb{SO}(3)$, and $\mathbf{p} = [\phi \mathbf{n}^T, \mathbf{t}^T]^T$. $[\mathbf{n}]\hat{\,}$ is the skew-symmetric matrix from $\mathbf{n}$, defined in [18]. $\mathbf{t} \in \mathbb{R}^3$ is the translation parameters.

**Connecting Shape with Image**  Given a projection function of the camera, we are able to project each point in the point cloud back to the image plane observed from a camera pose. However, one issue arises as we struggles to separate the truly visible points from those which are either self-occluded or out of image boundary. To circumvent this issue, traditional PBA methods either tie each 3D point to a pixel and attempt to estimate an inverse depth map [19], or allow 3 DoF for a point but only recovers the set of visible points from a particular frame [3]. In other words, historically there have been no attempts to install a full 3D point cloud for all views by chaining it with each frame. To deal with this issue, we take advantage of the recent advancement of "pseudo-render" [16], which is a differentiable projection function from a full 3D point cloud to a depth map given the camera pose. To achieve this, the "pseudo-render" projects the point cloud to an up-scaled plane and uses max pooling over the inverse depth channel to filter out invisible (occluded) points.

We write the "pseudo-render" as a function which outputs a subset $X_\mathbf{p}$ consisting of $M_\mathbf{p}$ visible points from the input point set $X$ of size $N$, given the camera pose $\mathbf{p}$:

$$X_\mathbf{p} := \{\widetilde{\mathbf{x}}_j\}_{j=1}^{M_\mathbf{p}} = \mathcal{PS}(X; \mathbf{p}). \quad (2)$$

**Photometric Consistency and The Optimization Policy**  Given an image sequence $\{\mathcal{I}_l\}_{l=0}^{L-1}$ and camera model, we write the optimization objective as the photometric loss between the reference frame and the subsequent frames, over point pairs, each consisting of one point visible in $\mathbf{p}_0$ and its corresponding point in target frame $l$ of pose $\mathbf{p}_l = \mathbf{p}_0 \circ \mathbf{\Delta p}_l$:

$$\mathcal{L}_{\text{ph}}(\mathbf{p}_0, \{\mathbf{\Delta p}_l\}_{l=1}^{L-1}, \mathbf{s}) = \quad (3)$$
$$\sum_{l=1}^{L-1} \sum_{j=1}^{M_{\mathbf{p}_0}} \Big(\mathcal{I}_0\big(\pi(\widetilde{\mathbf{x}}_j; \mathbf{p}_0)\big) - \mathcal{I}_l\big(\pi(\widetilde{\mathbf{x}}_j; \mathbf{p}_0 \circ \mathbf{\Delta p}_l)\big)\Big)^2$$

where $\{\widetilde{\mathbf{x}}_j\}_{j=1}^{M_{\mathbf{p}_0}} = \mathcal{PS}(\mathcal{G}(\mathbf{s}); \mathbf{p}_0)$, $M$ is figured out by the pseudo-renderer, dependent on $\mathbf{s}$ and $\mathbf{p}_0$. And the projection function with pose composition can be defined as:

$$\pi(\widetilde{\mathbf{x}}_j; \mathbf{p}_0 \circ \mathbf{\Delta p}_l) = \mathbf{K}(\mathbf{\Delta R}_l \mathbf{R}_0 \widetilde{\mathbf{x}}_j + \mathbf{R}_l \mathbf{t}_0 + \mathbf{\Delta t}_l) \quad (4)$$

assuming we write $\mathbf{p}_0 = [\phi_0 \mathbf{n}_0^T, \mathbf{t}_0^T]^T$, $\mathbf{\Delta p}_l = [\Delta \phi_l \mathbf{n}_l^T, \Delta \mathbf{t}_l^T]^T$, and $\mathbf{R}_0 = e^{[\mathbf{n}_0]\hat{\,}\phi_0}$, $\mathbf{\Delta R}_l = e^{[\mathbf{n}_l]\hat{\,}\Delta \phi_l}$.

The optimization policy of the photometric loss over all three set of variables is listed in the following Algorithm 1. We write the trained pose regressor as $\mathcal{R}_p(\mathcal{I}_0) = \mathbf{p}_0$ and style regressor as $\mathcal{R}_s(\mathcal{I}_0) = \mathbf{s}$, and we set the convergence threshold for $\mathcal{L}_{\text{ph}}$ to $\delta_L$.

---
**Algorithm 1** Optimization of the photometric loss
---
1: **procedure** $\mathcal{L}_{\text{PH}}(\mathbf{p}_0, \{\mathbf{\Delta p}_l\}_{l=1}^{L-1}, \mathbf{s})$
2:   $\mathbf{p}_0 \leftarrow \mathcal{R}_p(\mathcal{I}_0)$
3:   $\mathbf{s} \leftarrow \mathcal{R}_s(\mathcal{I}_0)$
4:   **while** $\mathcal{L}_{\text{ph}} > \delta_L$ or step < maximum iterations **do**
5:     **for** $l = 1, 2, \ldots, L$ **do**    ▷ in parallel
6:       $\mathbf{\Delta p}_l \leftarrow$ L-BFGS update on $\mathbf{\Delta p}_l$
7:     $\mathbf{p}_0 \leftarrow$ L-BFGS update on $\mathbf{p}_0$
8:     $\mathbf{s} \leftarrow$ L-BFGS update on $\mathbf{s}$
9:   **return** $\{\mathbf{p}_0, \{\mathbf{\Delta p}_l\}_{l=1}^{L-1}, \mathbf{s}\}$
---

It is worth mentioning that unlike traditional photometric bundle adjustment, we do not use Gauss-Newton optimization procedure here, because we are no longer able to calculate the warp Jacobian efficiently due to the non-linear and hidden relationship between style vector $\mathbf{s}$ and point cloud coordinates $\mathbf{x}_i$. Instead, we adopt a gradient-based optimization approach, *e.g.* SGD, LBFGS [17]. The gradients: $\partial \mathcal{L}/\partial \mathbf{s}$ and $\partial \mathcal{L}/\partial \mathbf{p}$ are efficiently calculated through back propagation with deep learning framework like Tensorflow [1].

In addition, it is interesting to note how the shape prior acts in the optimization process as compared with directly optimizing the set of visible points. Considering only a small fraction of $M$ points in the entire point cloud receives gradients because the rest do not contribute to the loss function visually, our approach using the shape prior is able to back propagate the gradients of the partial point cloud back to the style vector $\mathbf{s}$. After the update is made upon $\mathbf{s}$, a forward pass through the generator will change the positions of all points, including the invisible ones, thus molding both the visible and invisible parts. Moreover, changes are smoothed over neighbouring points owing to the convolutional point cloud generator. This mechanism can also be viewed as search through the learned style space. In contrast, in traditional PBA (e.g. Alismail *et al*. [3]) the position of each visible point is independently optimized, without a shape prior to chain them together, let alone recovering or constraining the invisible ones. We show visual comparisons between optimizing over the style vector and directly optimize the set of initialized points $X$ in Fig 7.

**Silhouette Constraints with 2D Chamfer Distance** In some cases, the photometric loss is not sufficient enough for constraining the shape, for example the optimization can plunge into a degenerate solution where all points are blown out of the frame, giving zero photometric loss, or the shape "shrink" to its center in an effort to avoid the large errors on surrounding edges. In this case, we propose to introduce another constraint, assuming we know the silhouette for the object in each frame (they can easily be acquired with modern instance segmentation methods).

For the $l_{th}$ frame, we acquire the set of pixel coordinates inside the silhouette $U_{l1} = \{\mathbf{u}_k \in \mathbb{R}^2\}_{k=1}^{K_{\mathbf{P}_l}}$. The set of projected visible point is then written as $U_{l2} = \{\mathbf{u}_j\}_{j=1}^{M_{\mathbf{P}_l}} = \{\pi(\widetilde{\mathbf{x}}_j)\}_{j=1}^{M_{\mathbf{P}_l}}$. Then the silhouette loss for all frames is the 2D Chamfer Distance between two sets of pixel locations:

$$\mathcal{L}_{\text{CD}}(\mathbf{p}_0, \{\mathbf{\Delta p}_l\}_{l=1}^{L-1}, \mathbf{u}) = \qquad (5)$$
$$\frac{1}{L-1} \sum_{l=1}^{L-1} \Big( \sum_{\mathbf{u}_k \in U_{l1}} \min_{\mathbf{u}_j \in U_{l2}} ||\mathbf{u}_k - \mathbf{u}_j||_2^2$$
$$+ \sum_{\mathbf{u}_j \in U_{l2}} \min_{\mathbf{u}_k \in U_{l1}} ||\mathbf{u}_j - \mathbf{u}_k||_2^2 \Big)$$

And the actual loss to optimize is weighted combination of $\mathcal{L}_{\text{ph}}$ and $\mathcal{L}_{\text{CD}}$:

$$\mathcal{L} = \mathcal{L}_{\text{ph}} + \lambda \mathcal{L}_{\text{CD}}, \qquad (6)$$

where $\lambda$ serves as the weighting parameter.

In the experiment, we report the performance of using photometric loss alone, as well as using combined loss.

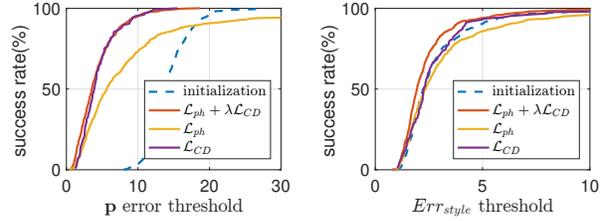

Figure 3: **Success rate of pose error and style error under varying thresholds.** We report results of three cases: (1) at initialization (dashed blue), (2) after optimization with photometric loss alone (solid yellow), and (3) after optimization with photometric loss and silhouette loss combined (solid red).

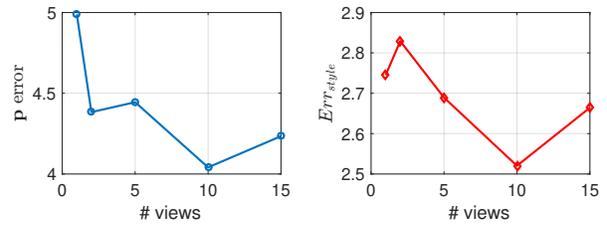

Figure 4: **Pose and style errors over different input sequence lengths.**

## 3. Experiments

### 3.1. Data Preparation

In evaluation, we utilize rendering engine Blender to synthesize object-centric RGB sequences with camera motions so that we are equipped with ground truth shape and pose for quantitative evaluation. We exploit the existing ShapeNet dataset [4] for a large collection of mesh-based CAD models in canonical pose for learning the shape prior of styles. Particularly, we select the 'car' category with 5,996 training and validation samples combined, and 1,500 samples for testing. For each aligned model, we uniformly sample 24,576 on the surface to give a dense point cloud.

In training the style and pose regressors, we apply camera motion to the scene. For each model, we render 200 samples with randomly sampled camera poses over the pose space, so in the end our training set consists of 1,199,200 image-shape pairs, with image resolution of 128×128. We also randomize the lighting conditions including environmental lighting by placing random number of lights with varying intensity and positions. Lambertian reflectance is imposed for better approximation to real-life statistics and the assumption of photometric consistency for points on the surface. An infinite textured carpet is laid on the ground to facilitate comparisons with feature-point-based traditional SfM methods (although a textured background is not required in our method considering we are using object-centric photometric loss).

|  | $Err_{\text{style}}$ | $\{\mathbf{p}_l\}$ mean error | $\mathcal{L}_{\text{ph}}$ | $\mathcal{L}_{\text{CD}}$ |
| --- | --- | --- | --- | --- |
| Zhu et al. [29] | 3.43 | 5.44 | - | $1.20\times 10^5$ |
| Ours (initialization) | 2.79 | 14.17 | 843.56 | $1.12\times 10^5$ |
| Ours (after optimization) | **2.70** | **4.24** | **197.37** | $\mathbf{7.64 \times 10^3}$ |
| Zhu et al. [29](1,536 points) | 3.82 | - | - | - |
| Fan et al.[7] (1,536 points) | 3.91 | - | - | - |

Table 1: **Quantitative evaluation on various metrics.** The error in $\mathbf{p}_0$ and $\mathbf{\Delta p}_l$ are reported in combination as mean error of $\{\mathbf{p}_l\}_{l=0}^{L-1}$ of all squences. Our method and single image reconstruction (Zhu et al. [29]) use a point cloud size of 24,576. Comparisons with [7] are carried out using a sparser target shape of 1,536 points.

To evaluate the optimization pipeline, we render one sequence for each model, with randomized starting pose $\mathbf{p}_0$, and then apply rotation to the camera around a random sampled axis to create a sequence of 15 frames, each of $2°$ rotation from its predecessor. In this case, each sequence covers camera motion of $30°$ in rotation.

### 3.2. Style and Pose Initialization

We first train a TL-embedding network with adapted architecture from Zhu et al. [29], including a style autoencoder with shapes in canonical poses, and a pose regressor plus a style regressor to map an input image with sampled pose and style to its ground truth. Before we begin to optimize the style and camera poses for an input sequence, we feed the first image (known as the reference frame) into the trained style and pose regressors to acquire a reasonable initialization for $\mathbf{s}$ and $\mathbf{p}_0$. The initial values of $\{\mathbf{\Delta p}_l\}_{l=1}^L$ are all set to zeros.

The initialization step can be viewed as single image style and pose prediction on the reference image. We give the quantitative results for all frames of all testing sequences in Table. 1, between two scenarios: 1) (row 1) predicting style and pose for all frames independently like Zhu et al. [29]; 2) (row 2) using initialized parameters based on the first frame. The loss in style $Err_{\text{style}}$ measures the 3D Chamfer Distance between the ground truth point cloud in canonical pose $X_{gt} = \{\mathbf{x}_{gt\ i}\}_{i=1}^N$ and the generated shape from the style vector $X = \{\mathbf{x}_i\}_{i=1}^N = \mathcal{G}(\mathbf{s})$:

$$Err_{\text{style}}(X_{gt}, X) = \qquad (7)$$
$$\frac{1}{N}\Big( \sum_{\mathbf{x}_{gt}\in X_{gt}} \min_{\mathbf{x}\in X} ||\mathbf{x}_{gt} - \mathbf{x}||_2^2 + \sum_{\mathbf{x}\in X} \min_{\mathbf{x}_{gt}\in X_{gt}} ||\mathbf{x} - \mathbf{x}_{gt}||_2^2 \Big)$$

To give readers an idea how our single-image-based style initialization performs as compared to the state-of-the-art, we also give a comparison with [7] on reconstructing a point cloud from each single frame of the sequences in the last two rows of Table. 1. To be consistent with [7] whose architecture only allows reconstruction of 1,536 points, we make a variant of our architecture (details in Appendix I), and retrain both methods on our synthetic test samples of varying style and pose, and measure their testing results w.r.t. $Err_{\text{style}}$. We show in Table. 1 that our architecture gives comparable performance in offering an initialization in style compared to [7], but we have demonstrated the ability to further optimize both style and poses through our pipeline and reach much finer results as also shown in Table. 1.

### 3.3. Quantitative Evaluation of the Optimization Pipeline

Our optimization pipeline is evaluated in three cases: optimize the photometric loss $\mathcal{L}_{\text{ph}}$, 2D Chamfer Distance $\mathcal{L}_{\text{CD}}$ separately, and optimize the combined loss $\mathcal{L} = \mathcal{L}_{\text{ph}} + \lambda\mathcal{L}_{\text{CD}}$. In our experiment we set the weight $\lambda$ to 0.01. We measure two metrics in Fig. 3 with success rate under threshold: the error in pose based on geodesic distance over the manifold of rotation [23](the smallest rotation angle from the ground truth pose to the estimated one), the error in style $Err_{\text{style}}$ between the generated point cloud in canonical pose with the sampled ground truth point cloud of the CAD model.

We also report the average performance at convergence in Table. 1 (row 3). The results are averaged over all frames in the testing sequences.

Another hyper parameter in our experiment is the number of views $L$ in a sequence. We show in Fig. 4 how does the number of views affects the metrics. We experiment with 2, 5, 10, 15 views, respectively covering a range of 4, 10, 20, 30 degree of camera rotation. One view corresponds to the single frame initialization. We conclude that as more observation taken into the optimization, both error generally decrease. An exception is the peak of style error in 2 views where the optimization get error-prone in face of extremely small camera motion. Also as number of views reach 15, the camera motion between the first and last frame becomes larger, and accordingly harder to optimize.

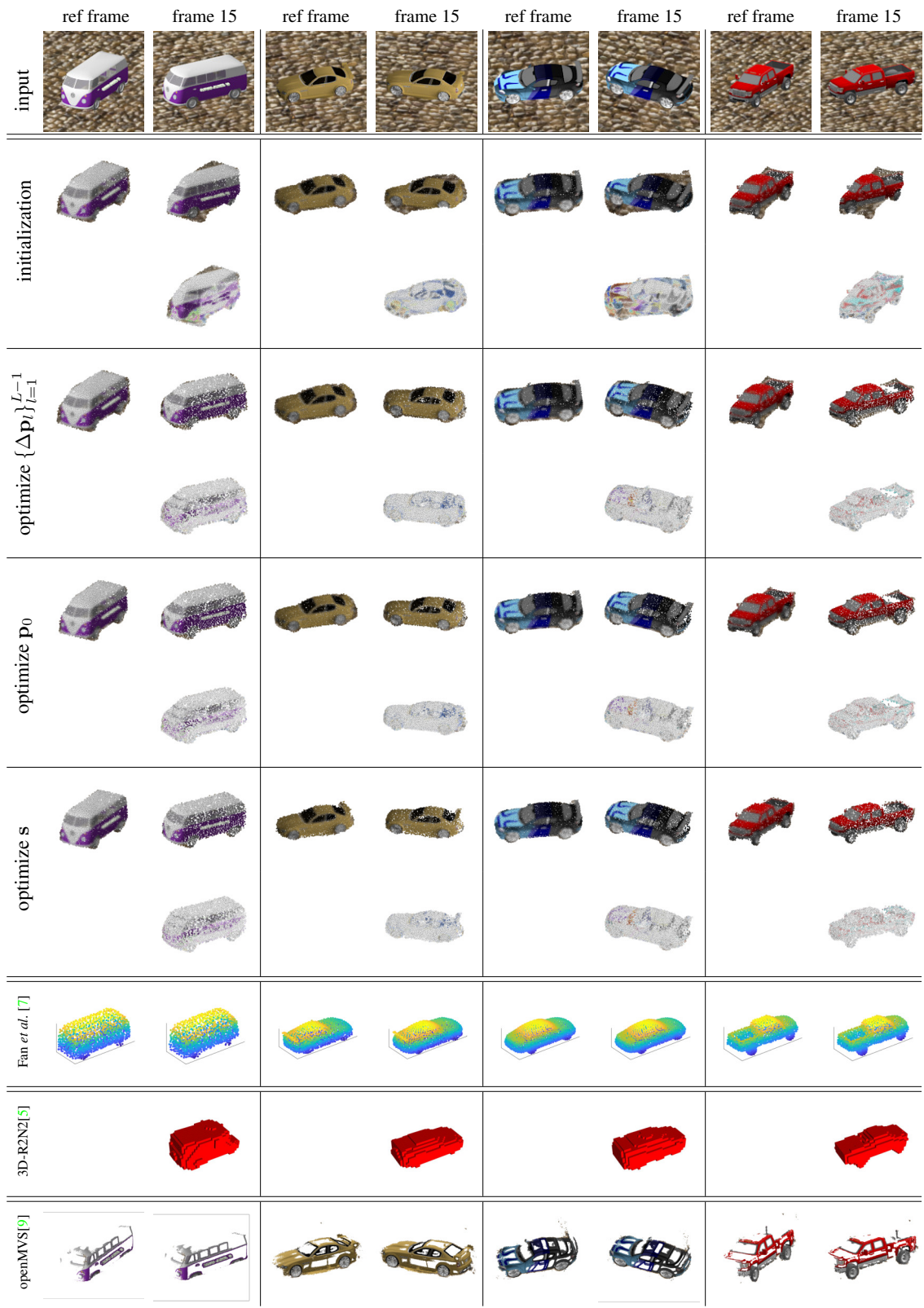

Figure 5: **Demo of our optimization pipeline on 4 sample sequences (only showing the reference and last frame), and comparisons with results from [7, 5, 9].** For each optimization step of each sample, we show in the cell the resampled points in the reference frame (upper left), resampled points from target frame (upper right), and the photometric residual between the former two (bottom right). We also give results from Fan *et al.* [7], 3D-R2N2 [5] and openMVS [9] in the last 3 rows.

## 3.4. Qualitative Comparisons and Analysis

We show in Fig. 5 step-by-step results of our optimization pipeline at initialization, and after each step of alternate optimization of poses and style. Take the first sequence of van for example: frame 15 aligns immediately back to the reference frame with a big jump after the optimization of $\Delta \mathbf{p}_{15}$. In the second step the pose of reference frame $\mathbf{p}_0$ is slightly adjusted for better alignment. In the end, $\mathbf{s}$ is improved to bring about a noticeable change in the shape of the van (e.g. the original flat head is adjusted to a square one to better fit the images).

We also give visual comparisons with Fan *et al.* [7], [5] and multi-view stereo method openMVS[9] on the samples. Particularly, we show reconstruction per frame in canonical camera for [7], the final output from the recurrent network in canonical pose after feeding in the entire sequence for [5], and the output shape under its estimated camera pose for [9]. During our evaluation with openMVS [9], openMVS tends to fail when there are not enough input frames. It breaks frequently with an input sequence of 15 frames so we use 45 frames instead. Also this method relies heavily on feature correspondence. Therefore we use high-resolution inputs of 512×512 to enable reasonable amount of feature point detected.

We observe (1) Fan *et al.* [7] gives pose-less shapes without much details, and consequently do not reproject back to the frames; (2) 3D-R2N2 [5] outputs voxelized shapes in canonical pose with error-prone styles; (2) openMVS fails on certain samples, while for the rest it produces sparse points at edges only, and it suffers from occasional problematic camera estimation. We are not able to benchmark with 3D-R2N2 or openMVS considering the difference in shape representation of the final product among those methods.

**Analysis on two options of losses.** In Fig. 6 we give an example of using three options of losses. As we may observe, using photometric loss can be problematic in that the optimizer gives lower photometric error by confidently cutting out some parts (parts in the windshield region in this case). With the additional constraint on silhouettes a better fit is reached.

**Analysis on using the shape prior.** As is shown in Fig. 7, if we directly optimize for the locations of visible points, the shape is noticeably degraded. However by optimizing the latent style vector, the style sticks around in the space of cars, generating a more smooth car-like shape.

## 4. Conclusion

In this paper we propose the method which combine geometric consistency and deep shape prior in the task of photometric bundle adjustment. More particularly, we devise the pipeline which takes in a full 3D point cloud implicitly parameterized by a style embedding, and jointly optimize the camera poses and the style vector based on photometric consistency and silhouette matching. We evaluate our method on object-centric video sequences both quantitatively and qualitatively, and offer comparisons with methods of SfM and shape-from-image(s). We are able to demonstrate our pipeline is capable of reconstructing an accurate dense point cloud as well as camera poses from video sequences of very few frames.

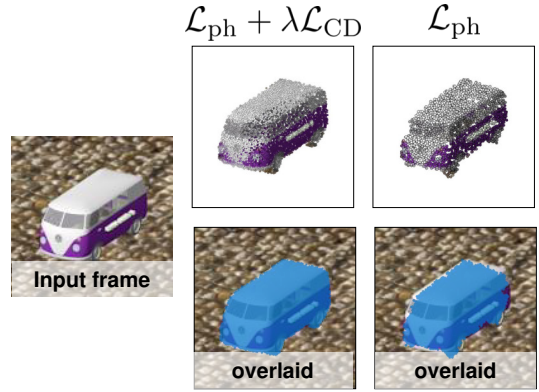

Figure 6: **Demo of using two options of loss in optimization.** The resampled points is overlaid in blue on the frame to demonstrate the silhouette matching.

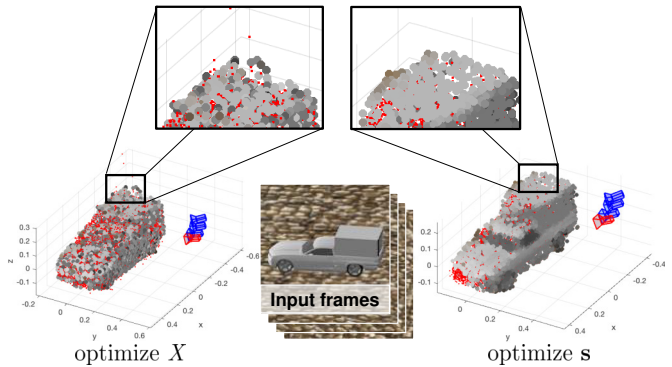

Figure 7: Demo of direct optimization ($\mathcal{L}_{\text{ph}} + \lambda \mathcal{L}_{\text{CD}}$) of point cloud positions (left) and optimization style through the shape prior (right).